\documentclass{article}

\usepackage{microtype}
\usepackage{graphicx}
\usepackage{subfigure}
\usepackage{booktabs}

\usepackage{hyperref}

\usepackage[accepted]{icml2019}

\usepackage{comment}

\usepackage{amsmath}

\usepackage{amssymb}
\newcommand{\mathbold}[1]{\ensuremath{\boldsymbol{\mathbf{#1}}}}

\usepackage[nameinlink]{cleveref}
\creflabelformat{equation}{#1#2#3}

\usepackage
[acronym,smallcaps,nowarn,section,nogroupskip,nonumberlist]{glossaries}
\glsdisablehyper{}

\DeclareRobustCommand{\mb}[1]{{\mathbold{#1}}}

\newcommand{\mbv}{\mb{v}}
\newcommand{\mbw}{\mb{w}}
\newcommand{\mbx}{\mb{x}}

\newcommand{\mbz}{\mb{z}}

\newcommand{\mbR}{\mb{R}}

\newcommand{\mbZ}{\mb{Z}}

\newcommand{\mbbeta}{\mb{\beta}}

\newcommand{\mbeta}{\mb{\eta}}

\newcommand{\mblambda}{\mb{\lambda}}
\newcommand{\mbmu}{\mb{\mu}}

\newcommand{\mbsigma}{\mb{\sigma}}

\newcommand{\mbtheta}{\mb{\theta}}

\newcommand{\mbvarepsilon}{\mb{\varepsilon}}

\DeclareMathOperator*{\argmax}{arg\,max}

\newcommand{\E}{\mathbb{E}}

\newcommand{\cL}{\mathcal{L}}

\newcommand{\cN}{\mathcal{N}}

\newcommand{\g}{\, | \,}

\usepackage{amsthm}

\newacronym{KL}{kl}{Kullback-Leibler}
\newacronym{ELBO}{elbo}{evidence lower bound}
\newacronym{EP}{ep}{expectation propagation}

\newacronym{MC}{mc}{Monte Carlo}
\newacronym{MCMC}{mcmc}{Markov chain Monte Carlo}

\newacronym{VI}{vi}{variational inference}
\newacronym{MFVI}{mfvi}{mean-field variational inference}
\newacronym{SVI}{svi}{stochastic variational inference}

\newacronym{VMP}{vmp}{variational message passing}

\newacronym{ADVI}{advi}{automatic differentiation variational inference}

\newacronym{RMSPROP}{rmsprop}{rmsprop}

\newacronym{NUTS}{nuts}{no-U-turn sampler}
\newacronym{HMC}{hmc}{Hamiltonian Monte Carlo}
\newacronym{VAE}{vae}{variational autoencoder}
\newacronym{VMF}{vmf}{variational matrix factorization}

\newacronym{ARD}{ard}{automatic relevance determination}
\newacronym{GMM}{gmm}{Gaussian mixture model}
\newacronym{DLGM}{DLGM}{deep latent Gaussian model}
\newacronym{LS}{ls}{Langevin-Stein}
\newacronym{OPVI}{opvi}{operator variational inference}

\newacronym{VPNG}{vpng}{variational predictive natural gradient}
\newacronym{NG}{ng}{natural gradient}

\icmltitlerunning{The Variational Predictive Natural Gradient}

\begin{document}

\twocolumn[
\icmltitle{The Variational Predictive Natural Gradient}

\icmlsetsymbol{equal}{*}

\begin{icmlauthorlist}
\icmlauthor{Da Tang}{columbia}
\icmlauthor{Rajesh Ranganath}{nyu}
\end{icmlauthorlist}

\icmlaffiliation{columbia}{Department of Computer Science, Columbia University, New York, New York, USA}
\icmlaffiliation{nyu}{The Courant Institute, New York University, New York, New York, USA}

\icmlcorrespondingauthor{Da Tang}{datang@cs.columbia.edu}
\icmlkeywords{Natural Gradient, Fisher Information, Variational Inference}

\vskip 0.3in
]

\printAffiliationsAndNotice{} 

\begin{abstract}
Variational inference transforms posterior inference into parametric optimization thereby enabling the use of latent variable models where otherwise impractical. However, variational inference can be finicky when different variational parameters control variables that are strongly correlated under the model. Traditional natural gradients based on the variational 
approximation fail to correct for correlations when the approximation is not the true posterior. To address this, we construct a new natural gradient called the Variational Predictive Natural Gradient (VPNG). Unlike traditional natural gradients for variational inference, this natural gradient accounts for the relationship between model parameters and variational parameters. We demonstrate the insight with a simple example as well as the empirical value on a classification task, a deep generative model of images, and probabilistic matrix factorization for recommendation.
\end{abstract}

\section{Introduction}
\label{sec:intro}

Variational inference \citep{jordan1999introduction} transforms
posterior inference in latent variable models into optimization. It
posits a parametric approximating family and tries to find the distribution in this family
that minimizes the \gls{KL} divergence to the posterior.
Variational inference makes posterior computation practical where it would not be otherwise.
It has powered many applications, including computational biology \citep{carbonetto2012scalable, stegle2010bayesian}, language \citep{miao2016neural},
compressive sensing \citep{shi2014correlated},
neuroscience \citep{manning2014topographic, harrison2010bayesian},
and medicine \citep{ranganath2016deep}.

Variational inference requires choosing an approximating family.
The variational family plus the model together define the
variational objective. The variational objective can be
optimized with stochastic gradients for a broad range of
models \citep{Kingma:2014,ranganath2014black,rezende2014stochastic}.
When the posterior has correlations, dimensions of the optimization
problem become tied, i.e., there is curvature. One way
to correct for curvature in optimization is to use natural
gradients \citep{amari1998natural, ollivier2011information, thomas2016energetic} .
Natural gradients for variational inference \citep{hoffman2013stochastic} adjust for the non-Euclidean nature of probability distributions. But they may not
change the gradient direction when the variational approximation is far
from the posterior.

To deal with curvature induced by dependent observation dimensions in the variational objective, we define a new type of natural gradient: the\glsreset{VPNG} \gls{VPNG}.
The \gls{VPNG} rescales the gradient with the inverse of the expected Fisher information matrix of the reparameterized model likelihood.
We relate this matrix to the negative Hessian of the expected log-likelihood part of the \gls{ELBO}, thereby showing it captures the curvature of variational inference.

Our new natural gradient captures potential pathological curvature introduced by the log-likelihood traditional natural gradient cannot capture. Further, unlike traditional natural gradients for variational inference, the \gls{VPNG} corrects for curvature
in the objective between model parameters and variational parameters.
In \Cref{sec:toy}, we will design an illustrate example where
the \gls{VPNG} points almost directly to the optimum, while both the vanilla gradient and the natural gradient point
in almost an orthogonal direction.

We show our approach outperforms vanilla gradient optimization and the traditional natural gradient optimization
on several latent variable models, including Bayesian logistic regression on synthetic data, variational autoencoders~\citep{Kingma:2014,rezende2014stochastic} on images, and variational probabilistic matrix factorization \citep{mnih2008probabilistic,gopalan2013scalable,liang2016modeling} on movie recommendation data.

\paragraph{Related work.}
Variational inference has been transformed by the use of Monte Carlo gradient estimators ~\citep{kingma2014adam,rezende2014stochastic,mnih2014neural,ranganath2014black,titsias2014doubly}. Though these approaches expand the applicability of variational inference, the underlying optimization problem can still be hard. Some recent work applied second-order optimization to solve this problem. For example, \citet{fan2015fast} derived Hessian-free style optimization for variational inference.
Another line of related work is on efficiently computing Fisher information and natural gradients for complex model likelihood such as the K-FAC approximation ~\citep{martens2015optimizing,grosse2016kronecker,ba2016distributed}. Finally, the \gls{VPNG} can be combined with methods for robustly setting step sizes, like using the \gls{VPNG} curvature matrix to build the quadratic approximation in TrustVI~\citep{regier2017fast}.

\section{Background}
\label{sec:bac}
\paragraph{Latent variable models}
Latent variable models posit latent structure $\mbz$ to
describe data $\mbx$ with parameters $\mbtheta$. The model is
\begin{align*}
p(\mbx, \mbz) = p(\mbz) p(\mbx \g \mbz ; \mbtheta).
\end{align*}
The model is split into a prior
over the hidden structure $p(\mbz)$ and likelihood
that describes the probability of data.
\paragraph{Variational inference} Variational inference~\citep{jordan1999introduction} approximates the posterior distribution $p(\mbz \g \mbx; \mbtheta)$ with
a distribution $q(\mbz \g \mbx; \mblambda)$ over the latent
variables indexed by parameter $\mblambda$. It works by
maximizing the \gls{ELBO}:
\begin{align}
\cL(\mblambda,\mbtheta) = \E_q\left[\log p(\mbx \g \mbz;\mbtheta)\right] - \textrm{KL}(q(\mbz  \g \mbx ; \mblambda) || p(\mbz))
\label{eq:elbo}
\end{align}
Maximizing the \gls{ELBO} minimizes the \gls{KL} divergence to the posterior. The model parameters $\mbtheta$ and variational parameters $\mblambda$ can
be optimized together. The
family $q$ is chosen to be amenable to stochastic optimization. One example is
the mean-field family, where $q(\mbz \g \mbx)$ is factorized over all coordinates of $\mbz$ like in the variational autoencoder.
\paragraph{$q$-Fisher information}
The \gls{ELBO} can be optimized with gradients. The effectiveness
of gradient ascent methods relates to the geometry of the problem.
When the loss landscape contains variables that control the objective
in a coupled manner, like the means of two correlated latent variables,
gradient ascent methods can be slow.

One way to adjust for this coupling or \emph{curvature} is to use
natural gradients~\citep{amari1998natural}. Natural gradients
account for the non-Euclidean geometry of parameters of probability
distributions by looking for optimal ascent directions in symmetric
KL-divergence balls.
The natural gradient relies on the Fisher information of $q$,
\begin{equation}
\begin{aligned}
F_q = \mathbb E_{q}\left[\nabla_\mblambda\log q(\mbz|\mbx;\mblambda)\cdot\nabla_\mblambda\log q(\mbz| \mbx; \mblambda)^\top\right].
\end{aligned}
\label{eq:q-fisher}
\end{equation}
We call this matrix the $q$-Fisher information matrix. With this Fisher information matrix, the \acrlong{NG} is
$\nabla_\mblambda^{\acrshort{NG}}\cL(\mblambda)~=~F_{q}^{-1}\cdot\nabla_\mblambda \cL(\mblambda)$.

Natural gradients have been used to optimize the the \gls{ELBO} \citep{hoffman2013stochastic}. The natural gradient works because it
approximates the Hessian of the \gls{ELBO} at the optimum.
The negative Hessian matrix of the \gls{ELBO} is:
\begin{equation}
\label{eq:q-fisher-derivation}
\begin{aligned}
-\frac{\partial^2}{\partial\mblambda^2} \cL = F_q+
&\int \frac{\partial^2}{\partial\mblambda^2}q \cdot (\log q(\mbz \g \mbx; \mblambda) - \log p(\mbz \g \mbx))d\mbz.
\end{aligned}
\end{equation}
The last integral in the above equation is small when the variational
distribution $q(\mbz \g \mbx; \mblambda)$ is close to the posterior distribution
$p(\mbz \g \mbx)$. Hence, the $q$-Fisher information matrix can be viewed as a
positive semi-definite version of the negative Hessian matrix of the \gls{ELBO}.
Thus natural
gradients improve optimization efficiency, when the variational approximation
is close to the posterior.

\section{The Variational Predictive Natural Gradient}
\paragraph{The $q$-Fisher information is insufficient.}
\label{sec:toy}

\begin{figure}
   \centerline{ \includegraphics[trim=40mm 10mm 40mm 8mm, width=0.5\columnwidth]{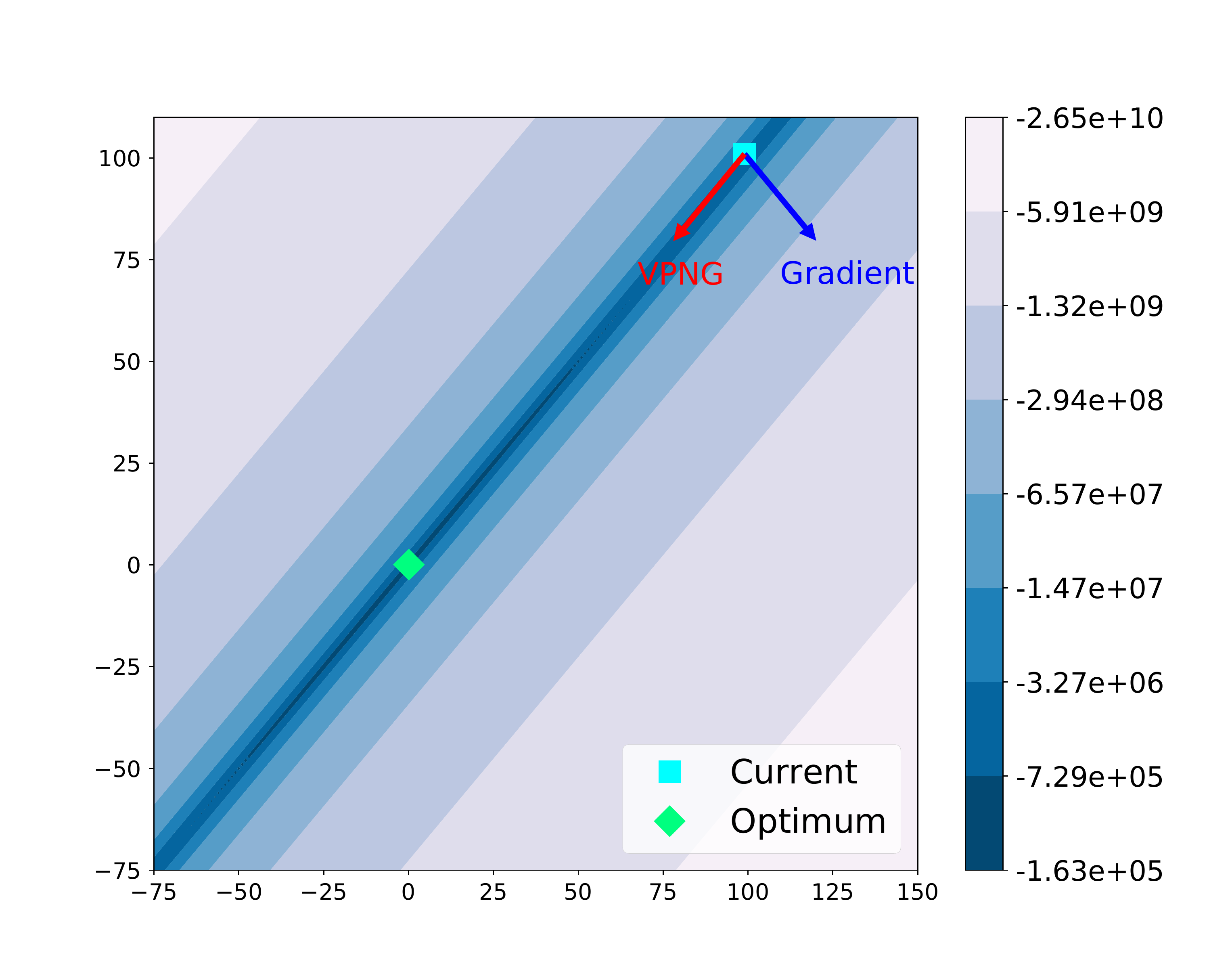}}
 \caption{
 The \acrshortpl{VPNG} are more effective than vanilla gradients and traditional natural gradients (pointing into the same direction with the vanilla gradients for this example).
\vspace{-.2in}
}
\label{fig:toy}
\end{figure}
Consider the following example with bivariate Gaussian likelihood that has an unknown mean $\mbmu=\begin{pmatrix}\mu_1\\\mu_2\end{pmatrix}$, a pathological known covariance $\Sigma=\begin{pmatrix} 1 & 1 - \varepsilon\\ 1-\varepsilon & 1\end{pmatrix}$ for some constant $0\le\varepsilon\ll 1$, and an isotropic Gaussian prior:
\begin{align}
p(\mbx_{1:n}, \mbmu) &=\ p(\mbmu \g \mb0, I_2)\prod_{i=1}^n \cN\left(\mbx_i \g \mbmu, \Sigma\right).
\label{eq:toy-model}
\end{align}
To do variational inference, we choose a mean-field approximation $q(\mbmu;\mblambda)=\cN(\mu_1\g\lambda_1,\sigma^2) \cN(\mu_2\g\lambda_2,\sigma^2)$ with $\sigma$ to be fixed. The posterior distribution for this problem is analytic: $p(\mbmu|\mbx)=\mathcal N(\mbmu',\Sigma')$ where $\Sigma'=(n\Sigma^{-1}+I_2)^{-1}$ and $\mbmu'=(n\cdot I_2+\Sigma)^{-1}\cdot\sum_{i=1}^n \mbx_i$.
The optimal solution for the variational parameter $\mblambda$ should be $\mbmu'$. The gradient of the objective function $\cL(\mblambda)$ is
\begin{align}
\nabla_\mblambda \cL(\mblambda)= -\mblambda + \Sigma^{-1}\cdot \left(-n\mblambda + \sum_{i=1}^n \mbx_i \right).
\label{eq:toy-gradient}
\end{align}
The precision matrix $\Sigma^{-1}$ is pathological. It has an eigenvector $\mbv_1=\frac1{\sqrt2}(1,1)^\top$ with eigenvalue $\frac{1}{2-\varepsilon}$, and an eigenvector $\mbv_2=\frac1{\sqrt2}(1,-1)^\top$ with eigenvalue $\frac{1}{\varepsilon}$. As a result, vanilla gradients will almost always
go along the direction of the eigenvector $\mbv_2$, as shown in \Cref{fig:toy}.
Further, natural gradients fail to resolve this. The $q$-Fisher information
matrix of this problem is diagonal, so it cannot help resolve the extreme curvature
between the parameters $\lambda_1$ and $\lambda_2$.

Notice that this pathological curvature is not due to that mean-field
approximation family on $q(\mbmu;\mblambda)$ does not contain the true posterior
$p(\mbmu \g \mbx)$. In fact, even if we optimize $q(\mbmu;\mblambda)$ over the
family of all bivariate Gaussian distributions $\cN(\mbmu \g
\mblambda_\mbmu,\mblambda_\Sigma)$, the partial gradient $\nabla_{\mblambda_\mbmu} \cL$ over the mean parameter vector $\mblambda_\mbmu$ will still have the same curvature issue. The issue arises since the variational approximation does not
approximate the posterior well at initialization  .
In general, if at some point the current $q$ iterate cannot approximate the posterior well, then the corresponding $q$-Fisher information matrix may not be able to correct the curvature in the parameters.

\subsection{Negative Hessian of the expected log-likelihood}
The pathology of the \gls{ELBO} for the model in \Cref{eq:toy-model} comes
from the ill-conditioned covariance matrix $\Sigma$.
The covariance matrix of the posterior can correct for this pathology
since its covariance matrix is $\Sigma'\approx\frac1n\Sigma$.
The disconnect lies in that variational inference is only
close to the posterior at its optimum, which implies that
q-natural gradients only correct for the curvature well
once the variational approximation is close to the
posterior, i.e., the inference problem is almost solved.

The problem is that the $q$-Fisher information matrix measures how parameter perturbations alter the variational approximation, regardless of the current model parameters and the quality of the current variational approximation.
We bring the model back into the picture by considering positive definite
matrices that resemble the
negative Hessian matrix of the expected log-likelihood part $\cL^{\textrm{ll}}=\E_q\left[\log p(\mbx \g \mbz;\mbtheta)\right]$ of the \gls{ELBO}, over both the variational parameter $\mblambda$ and the model parameter $\mbtheta$.

The expected log-likelihood contains where the
model and variational approximation interact, so
its Hessian contains the relevant curvature for
optimize the \gls{ELBO}. However,
since we are maximizing the \gls{ELBO}, the matrices need to not only
resemble the negative Hessian, but should also be positive semidefinite.
The negative Hessian of the expected log-likelihood is not guaranteed to be
positive semi-definite. Our goal is to construct a positive semidefinite
matrix related to the negative Hessian that accelerate inference by considering the curvature both the variational parameter and the model parameter.
In the sequel, we will show this new matrix is a type of Fisher information.

To compute gradients and Hessians, we need to compute derivatives
over expectations controlled by the variational parameter $\mblambda$.
In general, we can differentiate and use score function-style estimators
from
black box variational inference~\citep{ranganath2014black}. For simplicity,
consider the case where $q$ is reparameterizable~\citep{Kingma:2014, rezende2014stochastic}.
Then draws for $\mbz$ from $q$
can be written as deterministic transformations $g$ of noise terms $\mbvarepsilon$ with parameter-free distributions $s$. This simplifies the computations:
\begin{equation}
\mbz = g(\mbx,\mbvarepsilon;\mblambda)\sim q(\mbz \g \mbx;\mblambda)\iff\mbvarepsilon\sim s(\mbvarepsilon).
\label{eq:repara}
\end{equation}
The reparameterization trick can be applied to many common distributions (i.e. reparameterize a Gaussian draw $\nu~\sim~\cN(\mu,\sigma^2)$ as $\nu=\mu+\sigma\varepsilon$ where $\varepsilon\sim\cN(0,1)$).

With this trick, denote $\mbeta=(\mblambda^\top,\mbtheta^\top)^\top$, the negative Hessian matrix of $\cL^{\textrm{ll}}$ becomes:
\begin{equation*}
\begin{aligned}
&-\frac{\partial^2\cL^{\textrm{ll}}}{\partial\mbeta^2}=-\mathbb E_\mbvarepsilon\left[\frac{\partial^2}{\partial\mbeta^2}\log p(\mbx \g \mbz=g(\mbx, \mbvarepsilon;\mblambda);\mbtheta)\right].
\end{aligned}
\end{equation*}
Let us first consider the case where the variational distribution $q$ factorizes over data points: $q(\mbz \g \mbx; \mblambda)=\prod_{i=1}^nq(\mbz_i \g \mbx_i; \mblambda)$.  This factorization occurs in many popular models, such as in \glsreset{VAE}\glspl{VAE} \citep{Kingma:2014,rezende2014stochastic}.
Denote $Q$ as the empirical distribution of the observed data $\mbx_{1:n}$. Also denote $p(\mbz_i)$ and $p(\mbx_i \g \mbz_i)$ as the prior and likelihood function for any single data point $\mbx_i$. Moreover, for any data point $\mbx_i$ and $\mbx'_i$, we define the function
\[
u( \mbx_i,  \mbx'_i, \mbvarepsilon_i, \mbeta)=\frac{\partial^2}{\partial\mbeta^2}\log p( \mbx'_i \g  \mbz_i=g( \mbx_i,\mbvarepsilon_i; \mblambda); \mbtheta).
\]
Since we can use $\mbz_i=g(\mbx_i,\mbvarepsilon_i;\mblambda)$ to reparameterize $\mbz_i$, we can assume that the Jacobian matrix $\frac{\partial\mbz_i}{\partial\mbvarepsilon_i}$ is always invertible and hence by the \emph{inverse function theorem} we can also write $\mbvarepsilon_i$ as a function of $\mbz_i$, $\mbx_i$ and $\mblambda$. Hence, we can also express the above equation as
\[
\frac{\partial^2}{\partial\mbeta^2}\log p( \mbx'_i \g \mbz_i=g( \mbx_i,\mbvarepsilon_i; \mblambda); \mbtheta)=v( \mbx_i,  \mbx'_i, \mbz_i, \mbeta).
\]
With this notation, we can rewrite the above negative Hessian matrix for $\cL^{\textrm{ll}}$ as
\begin{equation}
\begin{aligned}
-&\frac{\partial^2\mathcal L^{\textrm{ll}}}{\partial\mbeta^2}
=-\sum\limits_{i=1}^n\mathbb \mathbb \E_{\mbvarepsilon_i}\left[\frac{\partial^2}{\partial\mbeta^2}\log p(\mbx_i \g  \mbz_i=g( \mbx_i,\mbvarepsilon_i; \mblambda); \mbtheta)\right]\\
&=-n\mathbb E_{Q(\mbx_i)}\left[\mathbb E_{\mbvarepsilon_i}\left[\frac{\partial^2}{\partial\mbeta^2}\log p(\mbx_i \g \mbz_i=g(\mbx_i,\mbvarepsilon_i; \mblambda); \mbtheta)\right]\right]\\
&=-n\mathbb E_{Q(\mbx_i)}\left[\mathbb E_{\mbvarepsilon_i}\left[u( \mbx_i, \mbx_i, \mbvarepsilon_i, \mbeta)\right]\right]\\
&=-n\mathbb E_{Q( \mbx_i)}\left[\mathbb E_{q( \mbz_i \g  \mbx_i;\mblambda)}\left[v( \mbx_i, \mbx_i, \mbz_i, \mbeta)\right]\right]\\
\end{aligned}
\label{eq:r-fisher0}
\end{equation}
Assessing the positive definiteness of \Cref{eq:r-fisher0} is a
challenge because of the expectation with respect to
the variational approximation. To make the positive definiteness easier
to wrangle, we make the assumption that
\begin{equation}
p(\mbz_i)p(\mbx_i \g \mbz_i)\approx Q(\mbx_i)q(\mbz_i \g \mbx_i).
\label{eq:dis-approx}
\end{equation}
When our model is learning a successful parameter vector $\mbeta$,
the distribution $p(\mbz_i,\mbx_i)~=~p(\mbz_i)p(\mbx_i \g \mbz_i)$
should be close to the distribution $Q(\mbz_i,\mbx_i)~=~Q(\mbx_i)q(\mbz_i \g \mbx_i)$ since the variational distribution $q$ is trying to learn the posterior distribution $p(\mbz_i \g \mbx_i)$
while $p(\mbx_i)$ is trying to learn the empirical data distribution $Q$. This
is the only approximation we will use to derive the \gls{VPNG}.

This substitution is similar to $q(\mbz \g \mbx)\approx p(\mbz \g \mbx)$ made
when analyzing the $q$-Fisher information matrix.
They can be quite different
when the $q(\mbz \g \mbx; \mblambda)$ approximating family may not be large enough to accurately approximate the posterior distribution $p(\mbz \g \mbx)$, and when the $p(\mbx \g \mbz; \mbtheta)$ model may not be able to accurately learn the data distribution $Q$.
With \Cref{eq:dis-approx} in hand, we have
\begin{align}
-&\frac{\partial^2\mathcal L^{\textrm{ll}}}{\partial\mbeta^2}
\approx-n\mathbb E_{p(\mbz_i)}\left[\mathbb E_{p( \mbx_i \g \mbz_i;\mbtheta)}\left[v(\mbx_i, \mbx_i, \mbz_i, \mbeta)\right]\right].
\label{eq:r-fisher1}
\end{align}
This matrix is computable via Monte Carlo, however in the next section we show
that this matrix may not be positive semidefinite and provide a method
to derive a matrix that is positive semidefinite.

\subsection{Predictive Sampling for Positive Semi-definiteness}
The inner expectation of \Cref{eq:r-fisher1} is an expectation of $v(\tilde \mbx_i, \tilde \mbx_i, \tilde \mbz_i, \mbeta)$ with respect to the distribution $p(\tilde \mbx_i \g \tilde \mbz_i;\mbtheta)$ on $\tilde\mbx_i$. This matrix appears to
be an average of Fisher information matrices, and thus positive semidefinite.
However, $v$ is not the Hessian of a distribution over $\mbx_i$ since $\mbx_i$
appears on both sides of conditioning bar. The failure of $v$ to be the Hessian
of a distribution for $\mbx_i$ means \Cref{eq:r-fisher1} may not be positive
definite. Next, we provide a concrete example where its not positive definite.

\paragraph{Non Positive Semi-definiteness of Second-Order Derivative.}
 Consider a model with data points $x_1,\ldots,x_n\in\mathbb R$ and local latent variables $z_1,\ldots,z_n\in\mathbb R$. The prior is $p(\mbz)=\prod_{i=1}^n\cN(z_i\g 0,1^2)$, the model distribution is $p(\mbx \g \mbz;\theta)=\prod_{i=1}^n\cN(x_i\g \theta z_i, 1^2)$ and the variational distribution is $q(\mbz \g \mbx; \lambda)=\prod_{i=1}^n\cN(z_i \g \lambda x_i, \sigma^2)$ with $\lambda,\theta\in\mathbb R$ and the hyperparameter $\sigma>0$. Then we can reparameterize each $\tilde z_i=\lambda\tilde x_i+\tilde\varepsilon_i$ with $\tilde\varepsilon_i\sim\cN(0, \sigma^2)$ drawn in an i.i.d. way. Under this model, \Cref{eq:r-fisher1} equals
\[
n\begin{pmatrix}
\theta^2(\theta^2+1) & \theta^2-1 \\
\theta^2 - 1 & 1
\end{pmatrix}\footnote{This matrix is normally related to both the variational parameter $\mblambda$ and the model parameter $\mbtheta$. Here this matrix is independent with $\mblambda$ since in this model $\frac{\partial \mbz}{\partial \mblambda}$ can be represented without $\mblambda$. The variational parameter will appear in this matrix if we set $z_i\sim\cN(\lambda^2x_i,\sigma^2)$ in this model.},
\]
which is not positive semi-definite when $|\theta|<\frac1{\sqrt3}$.

\paragraph{Predictive Sampling for Positive Semidefiniteness.}
The failure of the Hessian in \Cref{eq:r-fisher1} to be positive
definite stems from $v$ not being the Hessian of a probability
distribution. To remedy this, we sample the $\mbx_i$ on both side
of the conditioning bar independently. That is replace
\begin{equation}
\begin{aligned}
&\mathbb E_{p( \mbx_i \g \mbz_i;\mbtheta)}\left[v( \mbx_i, \mbx_i, \mbz_i, \mbeta)\right]\\
\end{aligned}
\end{equation}
with
\begin{align}
\mathbb E_{p(\mbx_i \g \mbz_i;\mbtheta)}\left[\mathbb E_{p(\mbx'_i \g \mbz_i;\mbtheta)}\left[v( \mbx_i, \mbx'_i, \mbz_i, \mbeta)\right]\right],
\label{eq:expect-approx}
\end{align}
where $\mbx'_i$ is a newly drawn data point from the same distribution $p(\cdot
\g \mbz_i;\mbtheta)$. This step is required. Rescaling the
gradient with the inverse of the first equation does not guarantee
convergence. This step will allows construction of a positive
definite matrix that captures the essence of the negative Hessian.
With this transformation, we get
\begin{equation}
\begin{aligned}
&-n\mathbb E_{p(\mbz_i)}\left[\mathbb E_{p(\mbx_i \g \mbz_i;\mbtheta)}\left[\mathbb E_{p( \mbx'_i \g \mbz_i;\mbtheta)}\left[v( \mbx_i,  \mbx'_i, \mbz_i, \mbeta)\right]\right]\right]\\
\approx&-n\mathbb E_{Q( \mbx_i)}\left[\mathbb E_{q( \mbz_i \g  \mbx_i;\mblambda)}\left[\mathbb E_{p( \mbx'_i \g \mbz_i;\mbtheta)}\left[v( \mbx_i, \mbx'_i, \mbz_i, \mbeta)\right]\right]\right]\\
=&n\mathbb E_{Q( \mbx_i)}\left[\mathbb E_{\mbvarepsilon_i}\left[\mathbb E_{p( \mbx'_i \g  \mbz_i=g( \mbx_i,\mbvarepsilon_i; \mblambda);\mbtheta)}\left[-u( \mbx_i,  \mbx'_i, \mbvarepsilon_i, \mbeta)\right]\right]\right].
\end{aligned}
\label{eq:r-fisher2}
\end{equation}
The approximation step follows from the earlier assumption that the joint
of $p$ and $q$ are close (see \Cref{eq:dis-approx}).

The inner expectation of the above equation is the negative Hessian matrix of the logarithm of the density of the distribution $p(\mbx'_i \g \mbz_i=g( \mbx_i,\mbvarepsilon_i; \mblambda);\mbtheta)$  with respect to the parameter $\mbeta$, given the latent variable $\mbvarepsilon_i$ and the data point $\mbx_i$. Therefore, this inner expectation equals the Fisher information matrix of this distribution, which is always positive semi-definite. The matrix in \Cref{eq:r-fisher2} meets our desiderata: it maintains structure from the negative Hessian of the
expected log-likelihood, is guaranteed to be positive semidefinite
for any model and variational approximation to that optimization converges,
and is computable via
Monte Carlo samples. To see that it is computable,

the matrix in \Cref{eq:r-fisher2} equals
\begin{equation}
\begin{aligned}
&n\mathbb E_{Q(\mbx_i)}\left[\mathbb E_{\mbvarepsilon_i}\left[\mathbb E_{p( \mbx'_i \g  \mbz_i=g( \mbx_i,\mbvarepsilon_i; \mblambda);\mbtheta)}\left[-u( \mbx_i, \mbx'_i, \mbvarepsilon_i, \mbeta)\right]\right]\right]\\
=&n\mathbb E_{Q( \mbx_i)}[\mathbb E_{\mbvarepsilon_i}[\mathbb E_{p( \mbx'_i \g  \mbz_i=g( \mbx_i,\mbvarepsilon_i; \mblambda);\mbtheta)}[\\
&\ \ \ \ \ \ \ \ \ \ \ \ \ \ \ \ \ \ \nabla_\mbeta\log p( \mbx'_i \g  \mbz_i=g( \mbx_i,\mbvarepsilon_i; \mblambda); \mbtheta))\\
&\ \ \ \ \ \ \ \ \ \ \ \ \ \ \ \ \ \ \cdot \nabla_\mbeta\log p( \mbx'_i \g \mbz_i=g( \mbx_i,\mbvarepsilon_i; \mblambda); \mbtheta))^\top]]].\\
\end{aligned}
\label{eq:r-fisher3}
\end{equation}
This equation can be computed by sampling a data point from the
observed data, sampling a noise term, and resampling a
new data point from the model likelihood.

\subsection{The variational predictive natural gradient}
The matrix in \Cref{eq:r-fisher3} is the expectation over a type of
Fisher information. First, define
\begin{equation*}
p(\mbx_i' \g  \mbz_i=g(\mbx_i,\mbvarepsilon_i; \mblambda); \mbtheta)
\label{eq:p-dis}
\end{equation*}
as the \emph{reparameterized predictive model distribution}.
The Fisher information of this matrix given $\mbx_i$
and $\mbvarepsilon_i$ is
\begin{align*}
F_{rep}(\mbx_i,\mbvarepsilon_i) =& \mathbb E_{p(\mbx_i' \g  \mbz_i=g(\mbx_i,\mbvarepsilon_i; \mblambda);\mbtheta)}[\\
&\nabla_\mbeta\log p(\mbx_i' \g  \mbz_i=g( \mbx_i,\mbvarepsilon_i; \mblambda); \mbtheta)\\
&\cdot \nabla_\mbeta\log p(\mbx_i' \g  \mbz_i=g(\mbx_i,\mbvarepsilon_i; \mblambda); \mbtheta)^\top].
\end{align*}
Averaging the Fisher information of the reparameterized predictive model distribution
over observed data points and draws from the variational approximation and rescaling
by the number of data points gives.
\begin{equation*}
\begin{aligned}
&n\mathbb E_{Q(\mbx_i)} \E_{\mbvarepsilon}[F_{rep}(\mbx_i,\mbvarepsilon_i)]\\
&=n\mathbb E_{Q( \mbx_i)}[\mathbb E_{\mbvarepsilon_i}[\mathbb E_{p( \mbx'_i \g  \mbz_i=g( \mbx_i,\mbvarepsilon_i; \mblambda);\mbtheta)}[\\
&\ \ \ \ \ \ \ \ \ \ \ \ \ \ \ \ \ \ \nabla_\mbeta\log p( \mbx'_i \g  \mbz_i=g( \mbx_i,\mbvarepsilon_i; \mblambda); \mbtheta))\\
&\ \ \ \ \ \ \ \ \ \ \ \ \ \ \ \ \ \ \cdot \nabla_\mbeta\log p( \mbx'_i \g \mbz_i=g( \mbx_i,\mbvarepsilon_i; \mblambda); \mbtheta))^\top]]]\\
&=\mathbb E_{\mbvarepsilon}[\mathbb E_{p(\mbx' \g  \mbz=g(\mbx,\mbvarepsilon; \mblambda);\mbtheta)}[\nabla_\mbeta\log p(\mbx' \g  \mbz=g( \mbx,\mbvarepsilon; \mblambda); \mbtheta)\\
&\ \ \ \ \ \ \ \ \ \ \ \ \ \ \ \ \ \ \cdot \nabla_\mbeta\log p(\mbx' \g  \mbz=g(\mbx,\mbvarepsilon; \mblambda); \mbtheta)^\top]]=: F_r.
\label{eq:r-fisher-info}
\end{aligned}
\end{equation*}
The positive semidefinite matrix related to
the negative Hessian of the \gls{ELBO} we derived in the previous
section is exactly the expected Fisher information of the
\emph{reparameterized predictive model distribution} $p(\mbx' \g  \mbz=g(\mbx,\mbvarepsilon; \mblambda);\mbtheta)$.

The expected density of reparameterized predictive model distribution can be viewed as the \emph{variational predictive distribution} $r(\mbx' \g \mbx; \mblambda, \mbtheta)$ of new data
\begin{equation*}
\begin{aligned}
\mathbb E_{\mbvarepsilon}\left[p(\mbx' \g  \mbz=g(\mbx,\mbvarepsilon; \mblambda); \mbtheta)\right]=&\mathbb E_{q(\mbz \g \mbx;\mblambda)}\left[p(\mbx' \g
 \mbz;\mbtheta)\right]\\
  :=&r(\mbx' \g \mbx; \mblambda, \mbtheta).
\end{aligned}
\end{equation*}
This distribution is the predictive distribution with the posterior replaced by the
variational approximation.
Hence, we call the matrix in \Cref{eq:r-fisher-info} as the \emph{variational predictive Fisher information} matrix.
This matrix can capture curvature. Though we derive it by assuming $q$ factorizes, this matrix may still capture curvature for the general case.

To illustrate that variational predictive Fisher information matrix can capture curvature, consider the example in \Cref{eq:toy-model}, we can reparameterize latent variable $\mbmu$ in the variational distribution $q(\mbmu;\mblambda)=\cN(\mu_1\g\lambda_1,\sigma^2) \cN(\mu_2\g\lambda_2,\sigma^2)$ as $\mbmu=\mblambda+\mbvarepsilon$, $\mbvarepsilon\sim\cN(\mb0, \sigma^2\cdot I_2)$.
Then the reparameterized predicted distribution $p(\mbx' \g \mbmu=\mblambda+\mbvarepsilon)$ equals $\prod\limits_{i=1}^n\cN(\mbx_i' \g \mblambda+\mbvarepsilon,\Sigma)$,
whose Fisher information matrix is just $n\Sigma^{-1}$. Hence the variational predictive Fisher information matrix for this model is $F_r=n\Sigma^{-1}$, which almost exactly matches with the pathological curvature structure in the gradient in \Cref{eq:toy-gradient}.

Therefore, our variational predictive Fisher information matrix contains the curvature we want to correct. Hence, we apply an update with the new natural gradient, the \emph{variational predictive natural gradient} (\gls{VPNG}):
\begin{align}
\nabla_{\mblambda,\mbtheta}^{\gls{VPNG}}\cL~=~F_r^{-1}\cdot\nabla_{\mblambda,\mbtheta} \cL(\mblambda,\mbtheta).
\label{eqn:vpng}
\end{align}
With this new natural gradient, the algorithm can move towards the true mean rather than getting stuck on the line $\lambda_1-\lambda_2=0$, as shown in \Cref{fig:toy}.

The variational predictive Fisher information matrix in \Cref{eq:r-fisher-info}
is a positive semi-definite matrix related to the negative Hessian of the
expected log-likelihood part $\cL^{\textrm{ll}}$ of the \gls{ELBO}. It can capture the curvature of variational inference since the expected log-likelihood part of the \gls{ELBO} usually plays a more important role in the whole objective and we can view the \gls{KL} divergence part $\textrm{KL}(q(\mbz \g \mbx) || p(\mbz))$ as a regularization for the $q$ distribution. In practice, the \gls{KL} divergence term gets scaled
by a ratio $\beta\in(0, 1)$ to learn better representations \citep{bowman2016generating}. With this scaling the curvature of the expected log-likelihood part $\cL^{\textrm{ll}}$ is even more important.

\subsection{Comparison with the traditional natural gradient}

The traditional natural gradient points to the steepest ascent direction of the \gls{ELBO} in the symmetric \gls{KL} divergence space of the variational distribution $q$
\citep{hoffman2013stochastic}. The \gls{VPNG} shares a similar type of geometric structure: it points to the steepest ascent direction of the \gls{ELBO} in the ``expected'' (over the parameter-free distribution $s(\mbvarepsilon)$ and data distribution $Q(\mbx)$) symmetric \gls{KL} divergence space of the reparameterized predictive distribution $p(\mbx' \g \mbz=g(\mbx, \mbvarepsilon; \mblambda); \mbtheta)$. Details are shown in appendix.

The $q$-Fisher information matrix tries to capture the curvature of the \gls{ELBO}. However, it strongly relies on quality of the fidelity of the variational approximation to the posterior, $q(\mbz \g \mbx)\approx p(\mbz \g \mbx)$.
The new Fisher information matrix, $F_r$ relies on a similar approximation $p(\mbz)p(\mbx \g \mbz)~\approx~ Q(\mbx)q(\mbz \g \mbx)$, these approximations are still quite different in many cases such as when the model does not
approximate the true data distribution well (described in the paragraph after \Cref{eq:dis-approx} ). Moreover, $F_r$ has the advantage that it considers the curvature from both the variational parameter $\mblambda$ and the model parameter $\mbtheta$ while the $q$-Fisher information matrix does not consider $\mbtheta$.

\section{Variational Inference with Approximate Curvature}

To build an algorithm with the \gls{VPNG}, we
need to compute the reparameterized predictive distribution and take an
expectation with respect to its Fisher information. These steps will only be tractable for
specific choices of models and variational approximations. We address
how to compute it with Monte Carlo in a broader setting here.

We can generate samples for $\mbx'$ in the distribution $p(\mbx' \g \mbz;  \mbtheta)$ for $\hat\mbz$ drawn from $q$. These samples can be used to estimate the
integrals in the definition of $F_r$. They are generated through
the following Monte Carlo sampling process. Using $k\in\{1,\ldots,M\}$ to index the Monte Carlo samples:
\begin{equation}
\hat\mbz\sim q(\mbz \g \mbx; \mblambda),\ \ \hat \mbx_i'^{(k)}\sim p(\mbx' \g \hat\mbz;\mbtheta).
\label{eq:r-approx}
\end{equation}
Reparameterization makes it easy to approximate the needed gradients of $\log p(\mbx_i'^{(k)} \g \hat\mbz ;  \mbtheta)$ with respect to $\mblambda$:
\begin{equation*}
\nabla_{\mblambda}\log p(\hat\mbx_i'^{(k)} \g \hat\mbz ;  \mbtheta)\approx\nabla_{\mblambda}\hat\mbz\cdot\nabla_{\hat\mbz_i}\log p( \hat\mbx_i'^{(k)} \g  \hat\mbz;  \mbtheta)^\top.
\end{equation*}
Denote $\hat b_{i,k}~=~\nabla_{\mblambda,\mbtheta}\log p(\hat\mbx_i'^{(k)} \g \hat\mbz;\mbtheta)$. Using samples from \Cref{eq:r-approx}, we can estimate the variational predictive Fisher information in \Cref{eq:r-fisher-info} as
\begin{equation}
F_r\approx \hat F_{r}=\frac1{M}\sum\limits_{k=1}^M\sum\limits_{i=1}^n\hat b_{i,k} \hat b_{i,k}^\top.
\label{eq:r-fisher-info-approx}
\end{equation}
This is an unbiased estimate of the variational predictive Fisher information matrix in \Cref{eq:r-fisher-info}.
\begin{algorithm}[tb]
   \caption{Variational inference with \acrshortpl{VPNG}}
  \label{alg:vpng-update}
\begin{algorithmic}
  \STATE {\bfseries Input:} Data $\mbx_{1:n}$, Model $p(\mbx, \mbz)$.
  \STATE Initialize the parameters $\mblambda$, and $\mbtheta$.
   \REPEAT
   \STATE Draw samples $\hat\mbz$ (\Cref{eq:r-approx}).
   \STATE Draw i.i.d samples $\hat \mbx_i'^{(k)}$ (\Cref{eq:r-approx}).
   \STATE Compute the Fisher information matrix $\hat F_r$ (\Cref{eq:r-fisher-info-approx}).
   \STATE Compute the natural gradient  $\hat\nabla_{\mblambda,\mbtheta}^{\gls{VPNG}}\cL$ (\Cref{eq:vpng-update}).
   \STATE Update the parameters $\mblambda$, $\mbtheta$ with the gradient $\hat\nabla_{\mblambda,\mbtheta}^{\gls{VPNG}}\cL$.
   \STATE (Optional) Adjust the dampening parameter $\mu$.
   \UNTIL{convergence}
\end{algorithmic}
\end{algorithm}

The approximate variational predictive Fisher information matrix $\hat F_r$ might be non-invertible. Since $\text{rank}(\hat F_r)\le Mn$, the matrix is non-invertible if $Mn<\text{dim}(\mblambda)+\text{dim}(\mbtheta)$. We add a small dampening parameter $\mu$ to ensure invertibility. This parameter can be fixed or dynamically adjusted. With this
dampening parameter, the approximate variational predictive natural gradient is
\begin{align}
\hat\nabla_{\mblambda,\mbtheta}^{\gls{VPNG}}\cL=(\hat F_r+\mu I)^{-1}\cdot\nabla_{\mblambda,\mbtheta}\cL.
\label{eq:vpng-update}
\end{align}
\Cref{alg:vpng-update} summarizes \gls{VPNG} updates. We set the dampening parameter $\mu$ to be a constant in our experiments. We show this algorithm works well in \Cref{sec:expe}.

\section{Experiments}
\label{sec:expe}
We explore the empirical performance of variational inference using the \gls{VPNG} updates in \Cref{alg:vpng-update}\footnote{Code is available at: \href{https://github.com/datang1992/VPNG}{https://github.com/datang1992/VPNG}.}.
We consider Bayesian Logistic regression on a synthetic dataset, the \gls{VAE} on a real handwritten digit dataset, and variational matrix factorization on a real movie recommendation dataset. We test their performances using different metrics on both train and held-out data.

We compare \gls{VPNG} with vanilla gradient optimization and traditional natural gradient optimization using 
RMSProp~\citep{tieleman2012lecture} and Adam~\citep{kingma2014adam} to set the learning rates in all three algorithms. For each algorithm in each task, we show the better result by applying these two learning rate adjustment techniques and select the best decay rate (if applicable) and step size.
We use ten Monte Carlo 
samples to estimate the \gls{ELBO}, its derivatives, and the 
variational predictive Fisher information matrix $F_r$.
\subsection{Bayesian Logistic regression}
\label{sec:blr}
We test \Cref{alg:vpng-update} with a Bayesian Logistic regression model on a synthetic dataset. We have the data $\mbx_{1:n}$ and the labels $y_{1:n}$ where $\mbx_i\in\mathbb R^4$ is a vector and $y_i\in\{0,1\}$ is a binary label. Each pair of $(\mbx_i,y_i)$ is generated through the following process:
\begin{equation*}
\begin{aligned}
a_i\sim&\text{Uniform}[-5,5]\in\mathbb R\\
\varepsilon_i^k\sim&\text{Uniform}[-0.005,0.005],\ k\in\{1,2,3,4\},\\
\mbx_i=&\left(a_i,\frac{a_i}2,\frac{a_i}3,\frac{a_i}4\right)+\mbvarepsilon_i\in\mathbb R^4,\\
y_i=&I\left[\langle(1,-2,-3,4), \mbx_i\rangle\ge0\right].
\end{aligned}
\end{equation*}
The generated data are all very close to the ground truth classification boundary $\langle(1,-2,-3,4),\mbx \rangle=~0$. We use Logistic regression with parameter $\mbw$ to model this data.
We place an isotropic  Gaussian prior distribution $p_0(\mbw)~=~\cN(\mbw\g \mb{0}, \sigma_0^2\cdot I_5)$ on the parameter $\mbw$ where the parameter $\sigma_0=100$. We apply mean-field variational inference to the parameter $\mbw$: $q(\mbw;\mbmu,\mbsigma)=\prod\limits_{i=1}^5\cN(w_i\g\mu_i,\sigma_i^2)$.  Mean-field variational families are popular primarily for their optimization efficiency. We aim to show that \gls{VPNG} improves upon the speed of mean-field approaches.
The data generative process and the initial prior parameter $\sigma_0$ makes the \gls{ELBO} pathological. Specifically, the covariates are strongly correlated while all data points have small margins with respect to the ground truth boundary.

We generate $500$ samples and select a fixed set which contains $80\%$ of the whole data for training and use the rest for testing. We test \Cref{alg:vpng-update} and the baseline methods on this data. We do not need Monte Carlo samples of predicted data
as the $F_r$ can be computed efficiently given samples from the latent variables in this problem.
To compare performances, we allow each algorithm to run 2000 iterations for 10 runs with various step sizes and compare the AUC scores for both the train and test procedure. The AUC scores are computed with the mean prediction.

The results are shown in \Cref{tab:auc}. In the experiments, we calculate the train and test AUC scores for every 100 iterations and and report the average of the last 5 outputs for each method.
\Cref{tab:auc} shows the train and test AUC scores for each method, over all 10 runs. Our method outperforms the baselines. We show a test AUC-iteration curve for this experiment in \Cref{sec:expe-appendix}. The vanilla gradient and traditional natural gradient do not perform well because of the curvature
 induced by the correlation in the covariates.
 \begin{table}[tb]
\caption{Bayesian Logistic regression AUC} \label{tab:auc}
\
\begin{center}
\begin{tabular}{lll}
\textbf{Method}  &\textbf{Train AUC} & \textbf{Test AUC} \\
\hline \\
Gradient         &$0.734\pm0.017$&  $0.718\pm0.022$\\
 \acrshort{NG}             &$0.744\pm0.043$ &$0.751\pm0.047$ \\
 \gls{VPNG}             &$\mb{0.972\pm0.011}$ & $\mb{0.967\pm0.011}$\\
\end{tabular}
\end{center}
\end{table}

\subsection{Variational autoencoder}
\label{sec:mnist}
We also study \glspl{VPNG} for \glsreset{VAE}\glspl{VAE} \citep{Kingma:2014,rezende2014stochastic} on binarized MNIST \citep{lecun1998gradient}. 
MNIST contains 70,000 images (60,000 for training and 10,000 for testing) of handwritten digits, each of size $28\times 28$.

We use a 100-dimensional latent representation $\mbz_i$. Our variational distribution
factorizes and we use a three-layer inference network to output the mean and variance
of the variational distribution given a datapoint. The generative model transforms $\mbz$ using
a three-layer neural network to output logits for each pixel. We use 200
hidden units for both the inference and generative networks.
 
To efficiently compute variational predictive Fisher information matrices, we view the entire \gls{VAE} structure as a 6-layer neural network
with a stochastic layer between the third and fourth layer. We then 
apply the tridiagonal block-wise Kronecker-factored curvature approximation (K-FAC), \citep{martens2015optimizing}. This enables us to compute Fisher information matrices faster in feed-forward neural networks.  We further improve efficiency by constructing low-rank approximations of large matrices. Finally, we use exponential moving
averages of quantities related to the K-FAC approximations. We show more details in appendix.
\begin{figure}[tb]
\vspace{-0.1in}
\centerline{
   \includegraphics[trim=60mm 10mm 70mm 0mm, width=0.75\columnwidth]{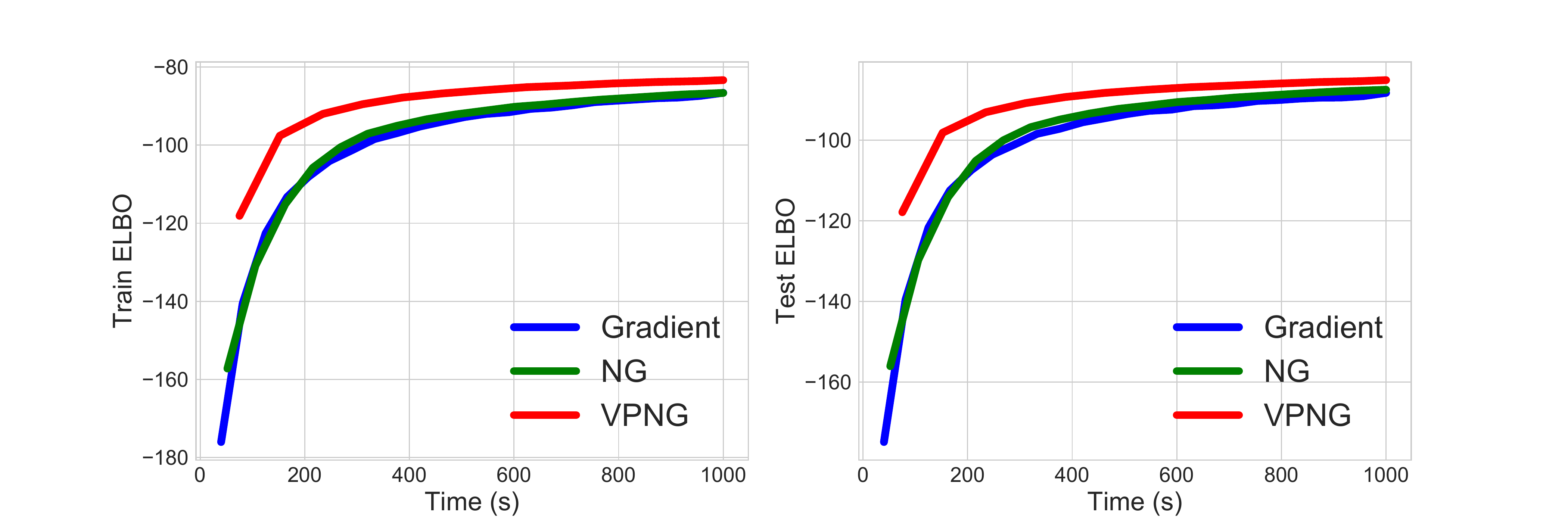}}
    \vspace{0.1in}
    \centerline{
    \includegraphics[trim=60mm 10mm 70mm 0mm, width=0.75\columnwidth]{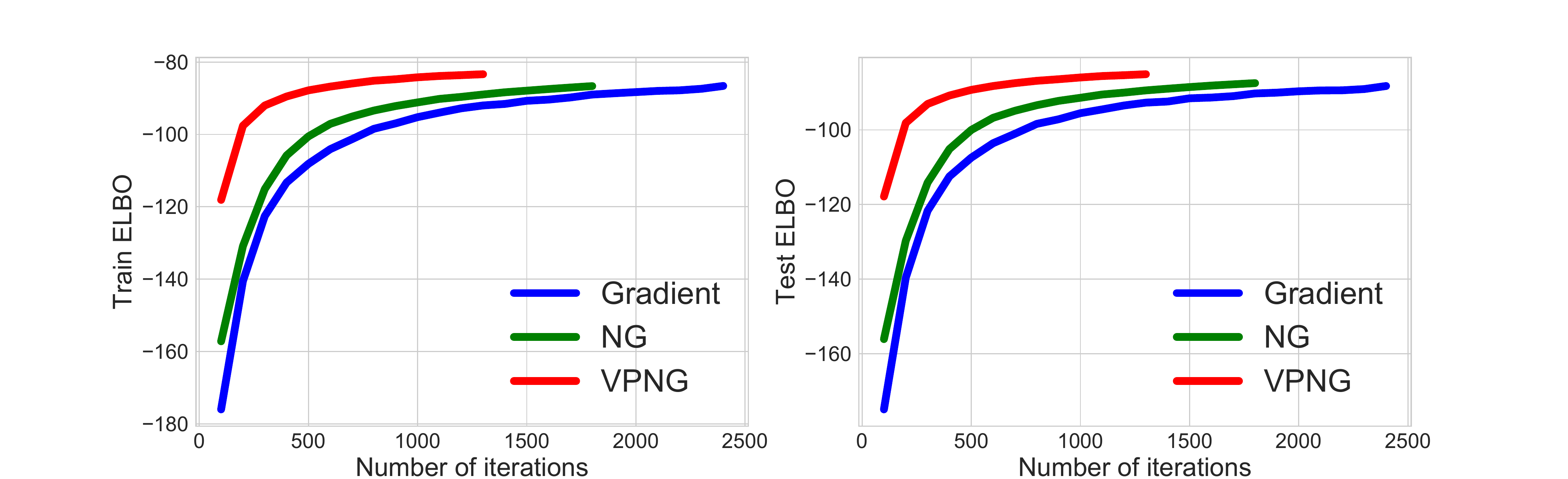}} 
   \caption{\gls{VAE} Learning curves on binarized MNIST}
\label{fig:bin-mnist}
\vspace{-0.15in}
\end{figure}

We compare the \gls{VPNG} method with the vanilla gradient and natural gradient optimizations. Since the traditional natural gradient does not deal with the model parameter $\mbtheta$, we use the vanilla gradient for $\mbtheta$ in this setting. We do not need to compare the performances of the \gls{VPNG} with the traditional natural gradient by fixing the model parameter $\mbtheta$ and learning only the variational parameter $\mblambda$ for two reasons. First, this setting is not common for \glspl{VAE}. Second, we need to have a fixed value for $\mbtheta$ and it is difficult to obtain an optimal value for it before running the algorithms.  

We select a batch size of 600 since we print the \gls{ELBO} values every 100 iterations. Hence, we evaluate the performances for each algorithm exactly once per epoch. The test \gls{ELBO} values are computed over the whole test set and the train \gls{ELBO} values are computed over a fixed set of 10,000 randomly-chosen (out of the whole 60,000) images. We allow each method to run for 1,000 seconds (we found similar results at longer runtimes) and select the best step sizes among several reasonable choices. \Cref{fig:bin-mnist} shows the results. Though the \gls{VPNG} method is the slowest per iteration, it outperforms the baseline optimizations on both the train and test sets, even on running time.  We also compare these methods with the second-order optimization method, the Hessian-free Stochastic Gaussian Variational Inference (HFSGVI)~\citep{fan2015fast}. However, it was not fast enough due to the large amount of Hessian-vector product computations. The \gls{ELBO} values with this method are still far below -200 within 1,000 seconds, which is much slower than the methods shown in \Cref{fig:bin-mnist}. 

The intuitive reason for the performance gain stems from the fact that the \gls{VAE} parameters
control pixels that are highly correlated across images. The \gls{VPNG}
corrects for this correlation.
\subsection{Variational matrix factorization}
\label{sec:movielens}
Our third experiment is on MovieLens 20M \citep{harper2016movielens}. This is a movie recommendation dataset that contains 20 million movie ratings from $n\approx135\text{K}$ users on $m_{\text{total}}\approx27\text{K}$ movies. Each rating $R^{\text{raw}}_{u,i}$ of the movie $i$ by the user $u$ is a value in the set $\{0.5,1.0,1.5,\ldots,5.0\}$. We convert
the ratings to integer values between 0 and 9 and select all movies with at least 5K ratings yielding $m\approx 1\text{K}$ movies. We model the zeros as in implicit matrix factorization \citep{gopalan2013scalable}.
We use Poisson matrix factorization to model this data. Assume there is a latent representation $\mbbeta_u\in\mathbb R^d$ for each user $u$ and there is a latent representation $\mbtheta_i\in\mathbb R^d$ for each movie $i$. Here $d=100$ is the latent variable dimensionality. Denote $\text{softplus}(t)=\log(1+e^t)$. We model the likelihood as
\begin{align*}
p(R\g \mbtheta,\mbbeta)=\prod\limits_{u=1}^n\prod\limits_{i=1}^m \text{Poisson}(R_{u,i}\g \mu=\text{softplus}(\mbbeta_u^\top\mbtheta_i)).
\label{eq:vmf-model}
\end{align*}
We do variational inference on the user latent variable $\mbbeta$ and treat the movie variables $\mbtheta$ as model parameters. The prior on each user latent variable is a standard Normal. We set the
variational distribution as $q(\mbbeta \g R; \mblambda)=\prod\limits_{u=1}^n q(\mbbeta_u \g \mbR_u; \mblambda)$, where $q(\mbbeta_u \g \mbR_u; \mblambda)$ uses an inference network that takes as input
the row $u$ of the rating matrix $R$. Similar to the \acrshort{VAE} experiment,
we use a 3-layer feed-forward neural network. We use 300 hidden units for this experiment.

Notice that the above likelihood is exactly a 1-layer feedforward neural network (without the bias term) that takes the latent representations drawn from the variational distribution $q(\mbbeta \g R; \mblambda)$ and outputs the rating matrix as a random matrix with a pointwise Poisson likelihood. Hence, we could view the model as a single-layer generative network and treat the latent variable $\mbtheta$ as its parameter. We have transformed variational matrix factorization to a task similar to the \gls{VAE}. Hence, when we apply \Cref{alg:vpng-update} to this model, we can apply the same tricks used in the \gls{VAE} experiments to accelerate the performances. We treat the whole model as a 4-layer feedforward neural network and again apply the tridiagonal block-wise K-FAC approximation \citep{martens2015optimizing} and adopt low-rank approximations of large matrices (again, more details in appendix).
\begin{figure}[tb]
\vspace{-0.1in}
\centerline{
   \includegraphics[trim=60mm 10mm 70mm 0mm, width=0.75\columnwidth]{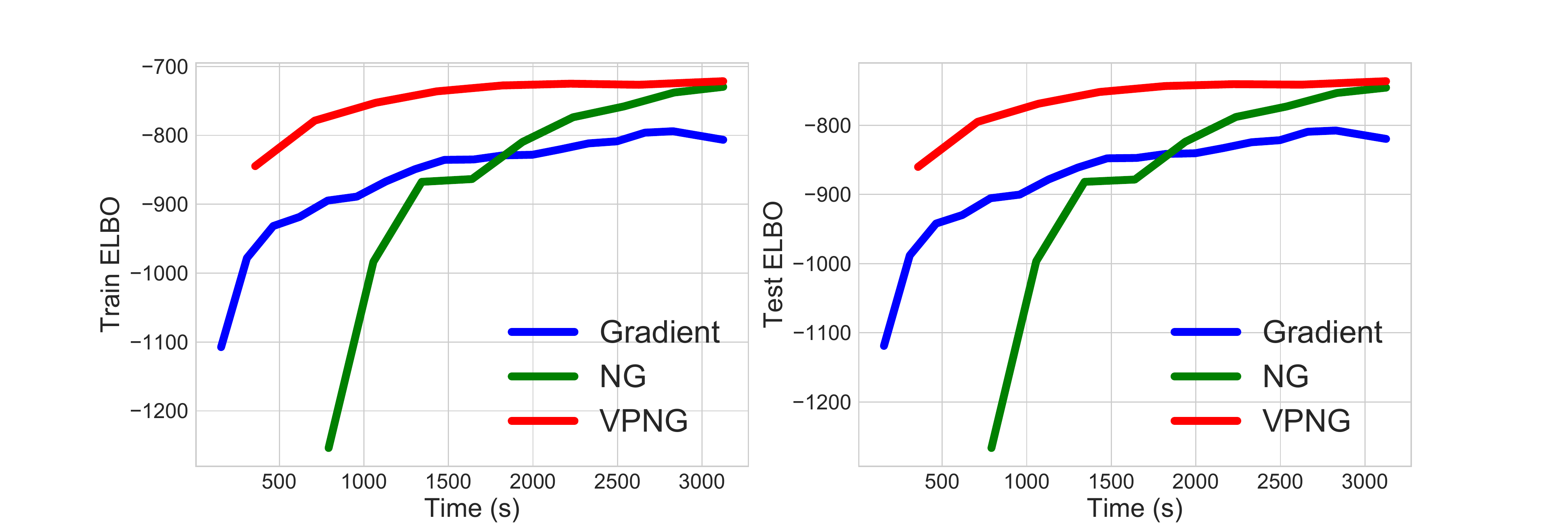}}
    \vspace{0.1in}
    \centerline{
    \includegraphics[trim=60mm 10mm 70mm 0mm, width=0.75\columnwidth]{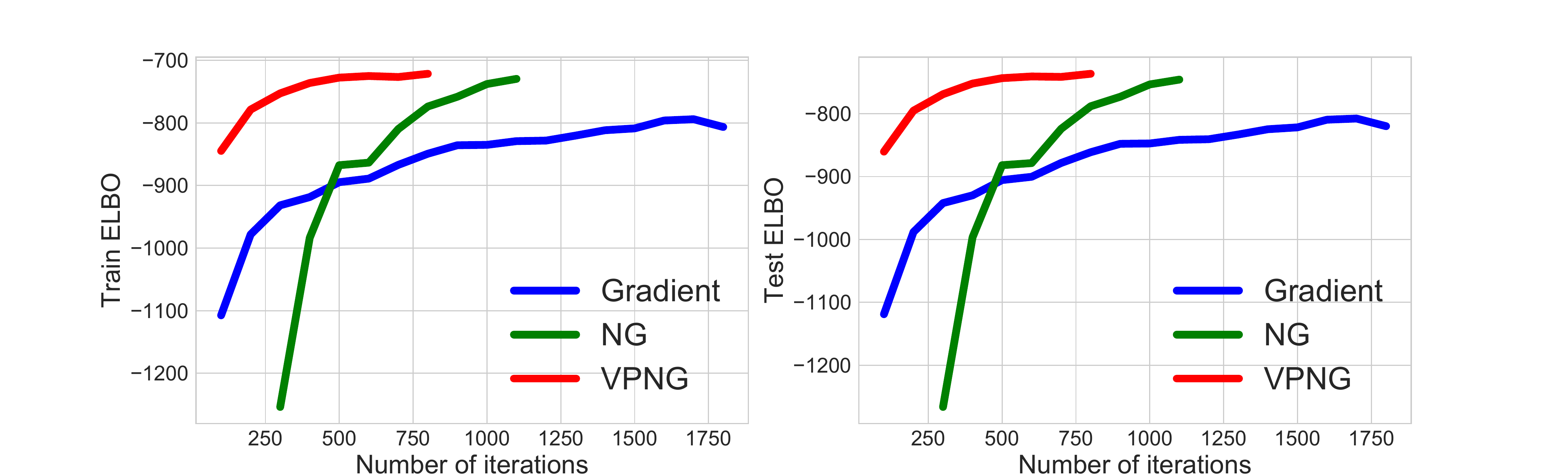}} 
 \caption{VMF Learning curves on MovieLens 20M}
\vspace{-0.15in}
\label{fig:vmf}
\end{figure}

The results are shown in \Cref{fig:vmf}. We randomly split the data matrix $R$ into train and test sets where the train set contains $90\%$ of the rows of $R$ (it contains ratings from $90\%$ of the users) and the test set contains the remaining rows. The test \gls{ELBO} values are computed over the random sampled test set and the train \gls{ELBO} values are computed over a fixed subset (with its size equal to the test set size) of the whole train set. Since this dataset is larger, we use a batch size of 3000. As can be seen in this figure, the \gls{VPNG} updates outperform the baseline optimizations on both the train and test learning curves. The curves look slightly different among various train/test splits of the dataset but \Cref{alg:vpng-update} consistently outperforms the baseline methods. The difference stems from the correlations in the ratings of the movies. The traditional natural gradient performs the worst at the beginning since it is only guaranteed to perform well at the end (when $q(\mbz \g \mbx)$ is close to the posterior distribution $p(\mbz \g \mbx)$, \Cref{eq:q-fisher-derivation} explains this), but not necessarily at the beginning, due to it does not consider potential curvature information in the model distribution. Across both experiments, we find that \gls{VPNG} dramatically improves estimation and inference at early iterations.

\section{CONCLUSION}
We introduced the variational predictive natural gradients. They adjust for parameter dependencies in variational inference induced by correlations
in the observations. We show how to approximate the Fisher information without manual model-specific computations. We demonstrate the insight on a bivariate Gaussian model and the empirical value on a classification model on synthetic data, a deep generative model of images, and matrix factorization for movie recommendation.
Future work includes extending to general Bayesian networks with multiple stochastic layers.

\section*{Acknowledgements}
\nocite{langley00}
We want to thank Jaan Altosaar, Bharat Srikishan, Dawen Liang and Scott Linderman for their helpful comments and suggestions on this paper.

\bibliography{paper}
\bibliographystyle{icml2019}

\newpage
\ 
\appendix
\newpage
\section*{Appendix}
We analyze the geometric structure of \gls{VPNG} to show its insight. Then we provide more details for the experiments.

\section{Analysis on the geometric structure of \acrshort{VPNG}}
\label{sec:geo-analysis}
As discussed in \citet{hoffman2013stochastic}, the traditional natural gradient points to the steepest ascent direction of the \gls{ELBO} in the symmetric \gls{KL} divergence space of the variational distribution $q$. Mathematically, for the \gls{ELBO} function as in \Cref{eq:elbo}, the traditional natural gradient points to the direction of the solution to the following optimization problem, as $\epsilon\rightarrow 0$:
\begin{equation*}
\begin{aligned}
\argmax_{\Delta\mblambda}&\ \ \ \ \cL(\mblambda+\Delta\mblambda,\mbtheta)\\
\text{s.t.}&\ \ \ \ \textrm{KL}_{\textrm{sym}}(q(\mblambda)||q(\mblambda+\Delta\mblambda))\le\epsilon.
\end{aligned}
\end{equation*}
In fact, denote $\mbeta=\begin{pmatrix}\mblambda \\ \mbtheta\end{pmatrix}$, the reparameterization for $\mbz=g(\mbx,\mbvarepsilon;\mblambda)$ and $p_{\mbx'}(\mbeta)=p(\mbx' \g \mbz=g(\mbx,\mbvarepsilon;\mblambda);\mbtheta)$ to be the reparameterized predictive distribution, our \gls{VPNG} (as defined in \Cref{eqn:vpng}) shares similar geometric structures and points to the direction of the solution to the following optimization problem, as $\epsilon\rightarrow 0$:
\begin{equation}
\begin{aligned}
\argmax_{\Delta\mbeta}&\ \ \ \ \cL(\mbeta+\Delta\mbeta)\\
\text{s.t.}&\ \ \ \ \mathbb E_\mbvarepsilon\left[\textrm{KL}_{\textrm{sym}}(p_{\mbx'} (\mbeta)||p_{\mbx'} ( \mbeta+\Delta\mbeta))\right]\le\epsilon.
\end{aligned}
\label{eqn:vpng-sym}
\end{equation}
Here the expectation on $\mbvarepsilon$ takes with respect to the parameter-free distribution $s(\mbvarepsilon)$ in the reparameterization.
\begin{proof}
The proof for the above fact is similar with the proof for the traditional natural gradient as in \citet{hoffman2013stochastic}. Ideally, we want to find a (possibly approximate) Riemannian metric $G(\mbeta)$ to capture the geometric structure of the expected symmetric \gls{KL} divergence $\mathbb E_\mbvarepsilon\left[\text{KL}_{\text{sym}}(p_{\mbx'} (\mbeta)||p_{\mbx'} (\mbeta+\Delta\mbeta))\right]$:
\begin{equation*}
\begin{aligned}
&\mathbb E_\mbvarepsilon\left[\text{KL}_{\text{sym}}(p_{\mbx'} (\mbeta)||p_{\mbx'} (\mbeta+\Delta\mbeta))\right]\approx\Delta\mbeta^\top G(\mbeta)\Delta\mbeta\\
&\ \ \ \ \ \ \ \ \ \ \ \ \ \ \ \ \ \ \ \ \ \ \ \ \ \ \ \ \ \ \ \ \ \ \ \ \ \ \ \ \ \ \ \ \ \ \ \ \ \ \ \ \ \ \ \ \ \ \ \ \ \ \ \ \ \ \ \ \ \ +o(\lVert\Delta\mbeta\rVert^2).
\end{aligned}
\end{equation*}
By making first-order Taylor approximation on $p_{\mbx'}(\mbeta+\Delta\mbeta)$ and $\log p_{\mbx'}(\mbeta+\Delta\mbeta)$, we get
\begin{equation*}
\begin{aligned}
&\mathbb E_\mbvarepsilon\left[\textrm{KL}_{\textrm{sym}}(p_{\mbx'} (\mbeta)||p_{\mbx'} (\mbeta+\Delta\mbeta))\right]\\
=&\mathbb E_\mbvarepsilon [\int (p_{\mbx'} (\mbeta+\Delta\mbeta)-p_{\mbx'} (\mbeta))\\
&\ \ \ \ \ \ \ \ \ \ \ \ \cdot(\log p_{\mbx'} (\mbeta+\Delta\mbeta)-\log p_{\mbx'} (\mbeta))d\mbx']\\
=&\mathbb E_\mbvarepsilon [\int (\nabla_\mbeta p_{\mbx'} (\mbeta)^\top\Delta\mbeta)\cdot(\nabla_\mbeta \log p_{\mbx'} (\mbeta)^\top\Delta\mbeta)d\mbx'\\
&\ \ \ \ \ \ \ \ \ \ \ \ +O(\lVert\Delta\mbeta\rVert^3)]\\
=&\mathbb E_\mbvarepsilon [\int p_{\mbx'} (\mbeta)\cdot(\nabla_\mbeta \log p_{\mbx'} (\mbeta)^\top\Delta\mbeta)\\
&\ \ \ \ \ \ \ \ \ \ \ \ \cdot(\nabla_\mbeta \log p_{\mbx'} (\mbeta)^\top\Delta\mbeta)d\mbx'+O(\lVert\Delta\mbeta\rVert^3)]\\
=&\Delta\mbeta^\top F_r\Delta\mbeta+O(\lVert\Delta\mbeta\rVert^3).
\end{aligned}
\end{equation*}
The term $O(\lVert\Delta\mbeta\rVert^3)$ is negligible compared to the first term when $\epsilon\rightarrow0$. Hence, we could take $G(\mbeta)$ to be just $F_r$, the variational predictive Fisher information as defined in \Cref{eq:r-fisher-info}. By \citet{amari1998natural}'s analysis on natural gradients, we know that the solution to \Cref{eqn:vpng-sym} points to the direction of $G(\mbeta)^{-1}\cdot \nabla_{\mblambda,\mbtheta}\cL=\nabla^{\gls{VPNG}}_{\mblambda,\mbtheta}\cL$, when $\epsilon\rightarrow0$.
\end{proof}

\section{More details for the experiments}
\label{sec:expe-appendix}
\begin{figure}[t]
\centerline{
    \includegraphics[trim=0mm 0mm 0mm 0mm, width=0.75\columnwidth]{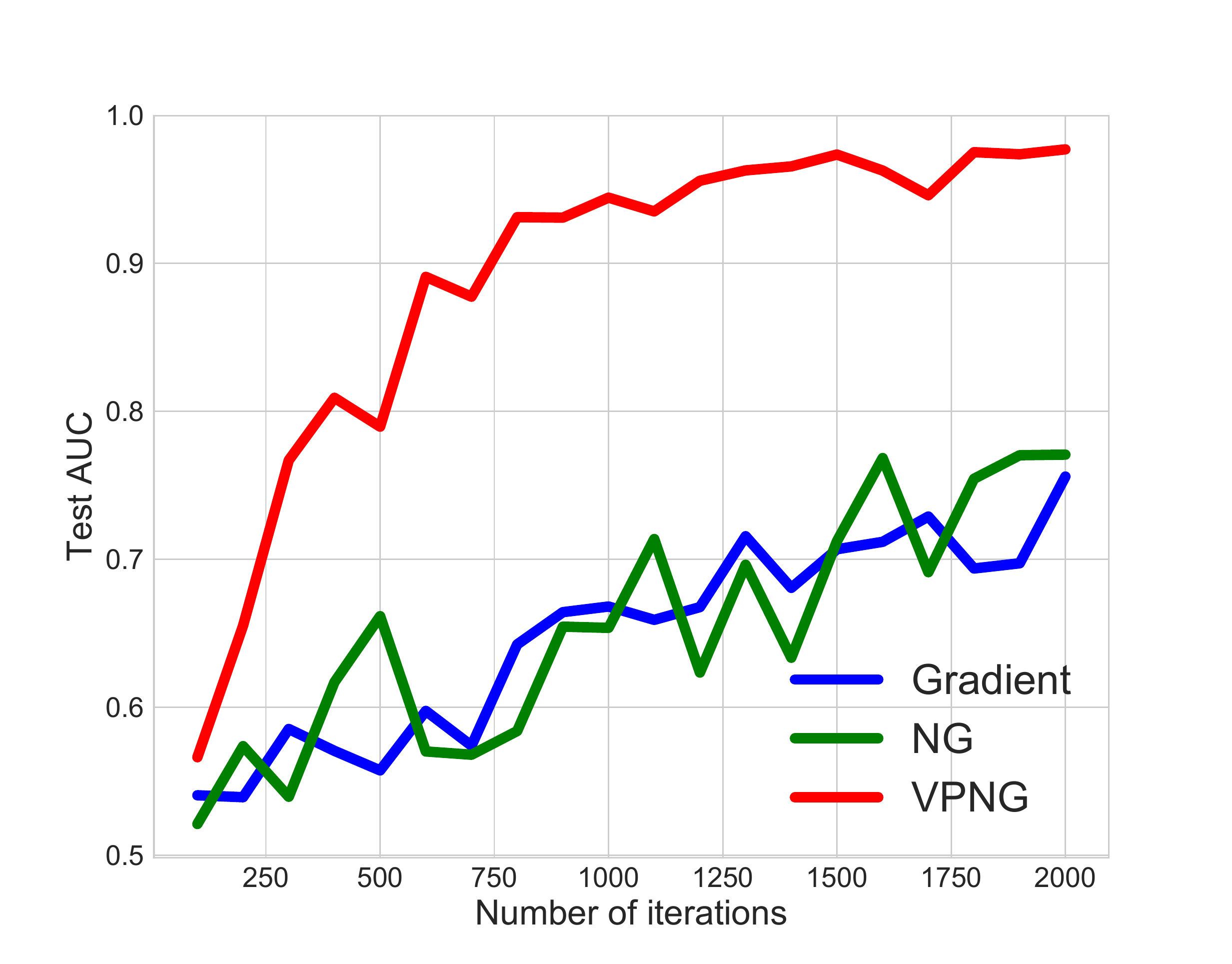}}
   \caption{Bayesian Logistic regression test AUC-iteration learning curve.}
\label{fig:auc-iter}
\vspace{-0.15in}
\end{figure}

For the Bayesian Logistic regression experiment, we show the test AUC-iteration curve as in \Cref{fig:auc-iter}. It can be seen that the \acrshort{VPNG} behaves more stable compared to the baseline methods.

For the \acrshort{VAE} and the \acrshort{VMF} experiments, we chose hyperparameters for all methods based on the training \gls{ELBO} at the end of the time budget.

For the traditional natural gradient and \gls{VPNG}, we applied the dampening factor $\mu$. More precisely, we take \gls{VPNG} updates as $\hat\nabla_{\mbtheta,\mblambda}^{\gls{VPNG}}\cL~=~(\hat F_r+\mu I)^{-1}\cdot\nabla_{\mbtheta,\mblambda}\cL$ and traditional natural gradient updates as $\hat\nabla_{\mblambda}^{\acrshort{NG}}\cL~=~(\hat F_q+\mu I)^{-1}\cdot\nabla_{\mblambda}\cL$). This is also applied in the Bayesian Logistic regression experiment in \Cref{sec:blr}. 

For the \acrshort{VAE} and the \acrshort{VMF} experiments, we applied the K-FAC approximation \citep{martens2015optimizing} to efficiently approximate the Fisher information matrices, in the \acrshort{NG} and \gls{VPNG} computations. For the \gls{VPNG}s, we view the \gls{VAE} model as a 6-layer neural network and the \acrshort{VMF} model as a 4-layer neural network. For the traditional \acrshort{NG}s, we view both the \acrshort{VAE} model and the \acrshort{VMF} model as 3-layer neural networks. We apply K-FAC on these models to efficiently approximate the Fisher information matrices with respect to the model distributions, given the samples from the variational distributions. 

To apply the K-FAC approximation, we will need to compute matrix multiplications and matrix inversions for some non-diagonal large square matrices (i.e. the $\bar A_{0,0}$ matrices in the K-FAC paper \citep{martens2015optimizing} and some other matrices that are computed during the K-FAC approximation process). In order to make the algorithms faster, we applied low-rank approximations for some large matrices of these forms by sparse eigenvalue decompositions. All of these large matrices are positive semi-definite. For each such large matrix $M$, we keep only the $K\cdot\ln(\text{dim}(M))$ dimensions of it with the largest eigenvalues and $K$ is a hyperparameter that can be tuned.

\begin{figure}[t]
\centerline{
   \includegraphics[trim=60mm 10mm 70mm 0mm, width=0.75\columnwidth]{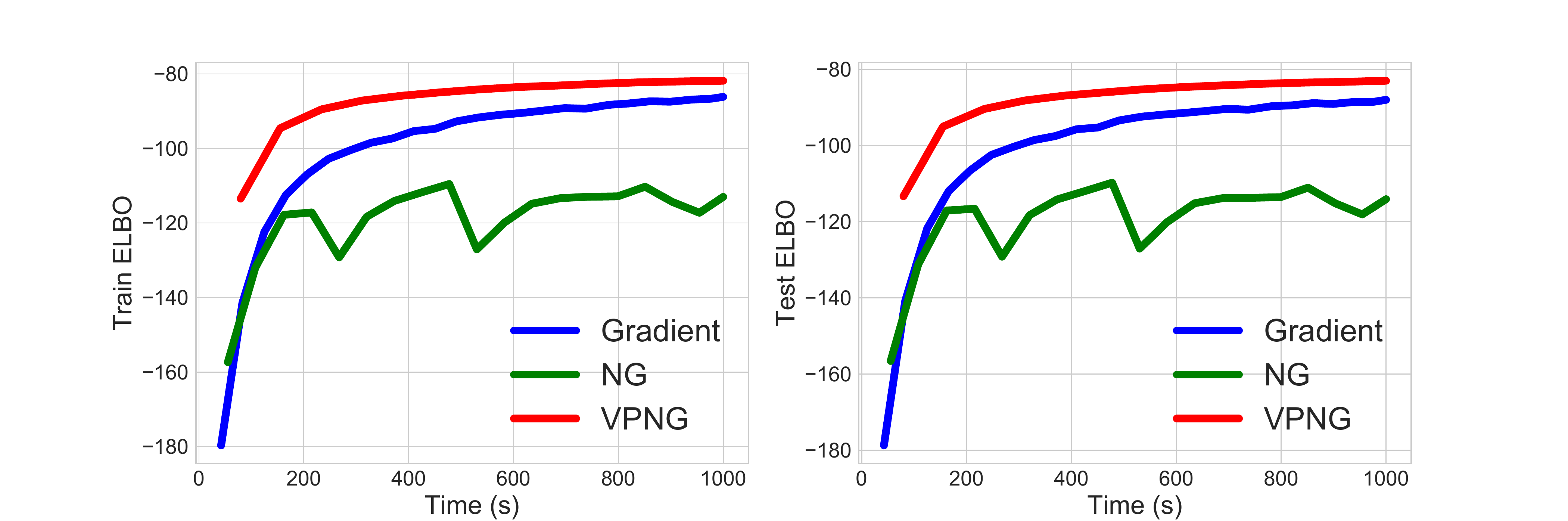}}
 \caption{\gls{VAE} learning curves on binarized MNIST, without exponential moving averages}
\label{fig:bin-mnist-no-ema}
\end{figure}

For \acrshort{NG} and \gls{VPNG}, we applied the exponential moving averages for all matrices $\bar A_{i,i}$ and $\bar G_{i+1,i+1}$ (again, we use the notations in the K-FAC paper \citep{martens2015optimizing}) to make the learning process more stable. We found that, by adding the exponential moving average technique, our \gls{VPNG} performs similarly to the case without this technique, while the traditional natural gradient is much more stable and efficient. If we do not apply the exponential moving average technique, the traditional natural gradients will not perform well. As an example, we show the performances of all methods without the exponential moving average technique in the \acrshort{VAE} experiment in \Cref{fig:bin-mnist-no-ema}.

We found that \acrshort{NG} and \gls{VPNG} performed similarly with respect to the dampening factor $\mu$, the exponential moving average decay parameter and the low-rank approximation function parameter $K$. However, different step sizes are needed to get the best performance from these two methods. We grid searched the step sizes and report the best one.

\end{document}